# An Efficient Scheduling for Security Constraint Unit Commitment Problem Via Modified Genetic Algorithm Based on Multicellular Organisms Mechanisms

Ali Yazdandoost, *Student Member, IEEE,* Peyman Khazaei, *Student Member, IEEE*, Rahim Kamali
*Student Member, IEEE*, Salar Saadatian, *Student Member, IEEE*

*Abstract*—Security Constraint Unit commitment (SCUC) is one of the significant challenges in operation of power grids which tries to regulate the status of the generation units (ON or OFF) and providing an efficient power dispatch within the grid. While many researches tried to address the SCUC challenges, it is a mixed-integer optimization problem that is difficult to reach global optimum. In this study, a novel modified genetic algorithm based on Multicellular Organisms Mechanisms (GAMOM) is developed to find an optimal solution for SCUC problem. The presentation of the GAMOM on the SCUC contain two sections, the GA and modified GAMOM sections. Hence, a set of population is considered for the SCUC problem. Next, an iterative process is used to obtain the greatest SCUC population. Indeed, the best population is selected so that the total operating cost is minimized and also all system and units' constraints are satisfied. The effectiveness of the proposed GAMOM algorithm is determined by the simulation studies which demonstrate the convergence speed. Finally, the proposed technique is compared with well-known existing approaches.

## I. INTRODUCTION

SECURITY Constraint Unit commitment (SCUC) is considered as a decision-making problem in power system operation issues. Indeed, based on the possible SCUC results, the day-ahead generators status determine to satisfy the load demand [1]. The goal of the SCUC issue is to supply the needed energy while the cost is minimized and all constraints associated with the problem are satisfied [1-3]. Hence, the output of the SCUC problem is ON or OFF status of generation units and also their hourly generation amount. Therefore, the status of generation units and hourly economic dispatch are two concerns that should be determine simultaneously.

Numerous evolutionary and mathematical techniques were investigated for SCUC problem, such as, mathematical modeling based on mixed-integer programming (MIP) [4], or evolutionary modeling based on Lagrangian relaxation (LR) [5], dynamic programing (DP) [6], particle swarm optimization (PSO) [7-9], and genetic algorithm (GA) [10]. The mentioned techniques have some benefit along with some disadvantages. It may be difficult to determine the global optimal solution with mathematical programming algorithms because SCUC is a mixed-integer non-convex optimization [11]-[12]. Due this drawback, evolutionary techniques have attracted a lots of attention to handle the SCUS problem. However, the effectiveness of the evolutionary techniques significantly related to their solution process. That means the solution can be vary on different techniques. Consequently, the performance along with the computational time are vary in the evolutionary techniques. Hence, in order to utilize the evolutionary technique, it is essential to select the algorithm carefully so that it can overcome the nonlinearity of the system.

Modified Genetic Algorithm Based on Multicellular Organisms Mechanisms (GAMOM) method is a novel evolutionary method [13-17]. It should be noted that, this algorithm has a sensible computation time and easy concept in comparison with the other existing evolutionary algorithms. Evolutionary algorithms are very popular in other engineering field as well [18-21]. Also, by combining them with other methods (such as fuzzy logic) can lead to an effective and high speed algorithm [22-25].

In this study, in order to solve the SCUC problem, the GAMOM method has been used. The main aim of this study is to reduce the total operating cost. It is worth to noting that in this paper, all the grid and generators constraints constraints are taken into consideration. Furthermore, spinning reserve is modeled to guarantee the satisfactory of operation under any

A. Yazdandoost Banizamani is with the Jahan Pardazesh Alborz Engineering Company Karaj – Iran ( e-mail: ali.yazdandoost7@gmail.com).

P. Khazaei is with the Department of Electrical and Computer Engineering, Shiraz University of Technology, Shiraz, Iran. (e-mail: electronic.peyman@gmail.com).

R. Kamali is with the Department of Electrical and Computer Engineering, Shiraz University of Technology, Shiraz, Iran. (e-mail: kamali.rahim@gmail.com).

S. Saadatian is with the Department of Mechanical, Louisiana State University, Baton Rouge, USA. (e-mail: ssaada1@lsu.edu).

load fluctuation [26-34]. It should be mentioned that, the reserve consideration can potentially increase the complexity of the SCUC problem and is difficult to be controlled by numerous existing evolutionary algorithms. To the best of the authors knowledge, this paper is the first paper that utilized the GAMOM algorithm in the power system operation problem. Results demonstrates the high superiority of the proposed algorithm.

## II. MATHEMATICAL MODELING

The main aim of the UC problem is to minimize the total generation cost. The total generation cost consists of the fuel and startup or shut down costs of the generating units [13-18].

$$minTC = \sum_{t \in \Omega_T} \sum_{i \in \Omega_g} f_i(p_{it}) I_{it} + SDU_{it} \quad (1)$$

where $TC$, $i$, $t$, $\Omega_g$ and $\Omega_T$ are the total cost, unit index, time, set of generating units and operating horizon respectively. It should be noted that the operating horizon is equal to 24 hours in this paper. The fuel startup and shutdown costs of unit $i$ is shown with $f_i(.)$ and $SUD_i$ respectively. The power generation of the unit $i$ in time $t$ is shown with $p_{it}$ and the binary variable which determines the ON/OFF status of unit $i$ at time $t$ is shown with $I_{it}$. The optimization problem is subjected to the following constraints.

### A. Power Balance
At any time, the total generated power should be equal to demand. That means:

$$\sum_{i \in \Omega_g} p_{it}.I_{it} \geq D_t \quad (2)$$

where $D_t$ is the total power demand at time $t$.

### B. Generating Units limitation:
This limitation can be defined as follow:

$$\underline{p_i}.I_{it} \leq p_{it} \leq \overline{p_i}.I_{it} \quad (3)$$

This constraint ensures that the power output of the generating unit $i$ is bounded by the unit's capacity. The minimum and maximum power generation of unit $i$ is shown with $\underline{p_i}$ and $\overline{p_i}$ respectively. It should be mentioned that, if the unit $i$ is OFF in the time period $t$, $I_{it}$ is equal to zero and thus $p_{it}$ is zero.

### C. Ramp up and down limitation
It is crucial to consider the ramping-up and ramping-down limits of unit $i$ while altering its power generation from time intervals $(t-1)$ to $t$. This assumption will lead to the following constraint[19-38]:

$$p_{it} - p_{i(t-1)} \leq RU_i + M(2 - I_{i(t-1)} - I_{it}) \quad (4)$$

$$p_{i(t-1)} - p_{it} \leq RD_i + M(2 - I_{i(t-1)} - I_{it}) \quad (5)$$

where $RU_i$ is the maximum ramp-up and $RD_i$ is the maximum ramp-down rate of unit $I$ in MW/hour, and $M$ is a large value.

### D. Startup and shutdown Constraints
Startup/shutdown prices of the generating units play a very significant role in the total operating costs. Two types of cost are determined in the UC problem, the hot start and cold start. When the unit is not cool-down totally and essentials to be turned on over it is called as the hot start. Also, the cold-start refers to the condition when the unit is cool-down completely. The startup cost can be expressed as follows:

$$St_i = \sum_{k=t-T_i^{off}-T_i^{cold}}^{t-1} U_i^k SU_i^t \quad (6)$$

Where

$$f(x) = \begin{cases} SU_i^{hot} U_i^t.(1-U_i^{t-1}), & \text{if } St_i^t > 0 \\ SU_i^{hot} U_i^t & \text{if } St_i^t = 0 \end{cases} \quad (7)$$

The shutdown cost equation can be determined with the same process.

### E. Spinning Reserve Constraints
Due to the unexpected disturbances, the power demand in time interval $t$ can be vary. Simultaneously, generating units may could not offer more power due to their limitations. Therefore, to prevent grid blackout and satisfying the require demand, sufficient reserve needs to be scheduled by the generation units. [13-18]. Although this constraint can lead to complexity of the problem, but it can significantly improve the reliability and security of the grid. In this paper, to consider the spinning reserve, the following constraint is expressed as follows:

$$\begin{cases} SR_{up}^t = \sum_{i=1}^{N} \min(P_{i,\max} - P_i^t, M \times r_i) U_i^t \\ SR_{up}^t = \sum_{i=1}^{N} \min(P_i^t - P_{i,\min}^t, M \times r_i) U_i^t \end{cases} \quad (8)$$

The spinning reserve requirements to support the unpredicted load increase or generation outrage is shown with $SR_{up}$ and the required spinning reserve to respond to the load reduction is shown with $SR_{dn}$. Also, the coefficient of the ramp rate is shown with $r$ which is equal to 10, which means the power output of the generating unit can alter within 10 minutes' time interval.

### F. Voltage and Angle constraints
The voltage and angle of each feeder should be within a limit in any time as (9) and (10).

$$\underline{V} \leq V_{m,t} \leq \overline{V} \quad (9)$$

$$\underline{\theta} \leq \theta_{m,t} \leq \overline{\theta} \quad (10)$$

### G. Power flow constraint:
The power flow of lines should be within the limit as (11).

$$S_{nm,t} \leq \overline{S} \quad (11)$$

where n and m are feeder's nodes, and *m* represents as the feeder index.

## Ⅲ. Proposed Genetic Algorithm based on Multicellular Organisms Mechanisms

The proposed GA follows mitosis and meiosis in the human cell. Previously, conventional GAs mimicked the meiosis processes on sexual chromosomes. In the proposed GA, the concept of mitosis for the asexual chromosomes will be added to GA. The flowchart for the proposed GA is illustrated in Fig.2. The proposed GA starts by producing chromosomes for the populations (population1 and population2) which are shown with $N_1$ and $N_2$ with the size of $n_1$ and $n_2$. Chromosome generation is a complete random process, and the chromosomes for each population are produced independently.

Table I
General algorithm's parameters

| Parameter | Definition | Number |
|---|---|---|
| $m_1$ | Mitosis rate | [0,1] |
| $m_2$ | Meiosis rate | [0,1] |
| $\alpha_1.m_1$ | Ratio of chromosomes in $N_1$ that should place in $N_1$' | [0,1] |
| $\alpha_2.m_2$ | Ratio of chromosomes in $N_2$ that should place in $N_2$' | [0,1] |
| $\beta_1.m_1$ | Ratio of chromosomes in $N_2$ that should place in $N_1$' | [0,1] |
| $\beta_2.m_2$ | Ratio of chromosomes in $N_1$ that should place in $N_2$' | [0,1] |

The parameters are set after generating populations, The definition of parameters such as $m_1$, $m_2$, $\alpha_1$, $\alpha_2$, $\beta_1$, $\beta_2$, and their range is introduced in Table 1. The parameters are important as they mix the chromosomes from population1 and population2 to make sub-population1 and sub-population2.

The two subpopulations are shown as $N_1$ and $N_2$. The sizes of subpopulations are formulated as follows:

$$n'_1 = \alpha_1.m_1.n_1 + \beta_1.m_1.n_2 \quad (12)$$

$$n'_2 = \alpha_2.m_2.n_1 + \beta_2.m_2.n_2 \quad (13)$$

The aim of the parameters is to determine how the algorithm collects sexual chromosomes and sexual chromosomes in to the subpopulation1 and subpopulation2 for mitosis and meiosis purposes.

A group of asexual chromosomes in $N_1$ undergoes mitosis, which consists of duplication and mutation actions. At the same time, the $N_2$ which consists of sexual chromosomes undergoes homologous recombination or crossover in meiosis phase I. There is a possibility of mutation in meiosis phase II. Normally, there is not a high rate of mutation in the mitosis and meiosis processes.

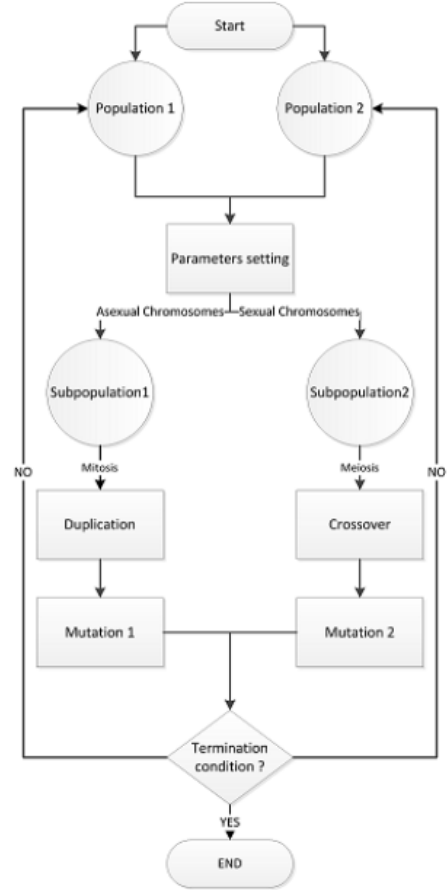

Fig .1. Flowchart of the proposed GA

In genetics, there are various factors for example exogenous (environmental factors) and endogenous (errors during DNA replications) which may be the reason for mutations. Mutations happen randomly. The randomness in mutations may have either no effect, alter the product of a gene, or prevent the gene from functioning appropriately. In the proposed algorithm, mutations have similar concept of randomness but the mutations in mitosis and meiosis are in different forms. The rate of mutations in mitosis is selected to be higher than meiosis to follow the natural mechanisms of cell division. At the end, after mutation in mitosis and meiosis, the successive chromosomes in mitosis are going to $N_1$. Furthermore, the chromosomes at the end of meiosis are going to $N_2$ based on elite characteristics of the chromosomes. Both mitosis and meiosis execute at the same time in parallel.

It should be mentioned that, the proposed method has an advantage over the other methods which is the migration of the chromosomes between the subpopulations. However, it has a significant dissimilarity between proposed model and Island model which has been introduced in the literature. In Island model, all subpopulations are searching for the optimum solution and sometimes the island with disfavored results die and the island with the superior results survive. However, in this model none of the subpopulations are going to be deactivated. It is a mechanism to keep the variety in the population. Fig. 1 show the flowchart of the proposed algorithm.

## IV. Simulation Result

In order to validates the effectiveness of the proposed algorithm, two different cases are considered as follows:

*Case1: Solving the US by considering three generation units.*

*Case2: Solving the SCUS by considering ten generation units.*

**Case1:** In this case a small networked is considered which includes three generation units. Table II describes the characteristics of distributed generators (DGs). Fig.2 demonstrates the required energy for the day-ahead market.

Table II
Characteristics of DGs

| Type | Min-Max Capacity (kW) | Cost ($/kWh) | Up/Down | Min Up/down Time (h) |
|---|---|---|---|---|
| DG1 | 20-100 | 0.1 | 100 | 3 |
| DG2 | 40-50 | 0.2 | 40 | 3 |
| DG3 | 1-25 | 0.4 | 0 | 3 |

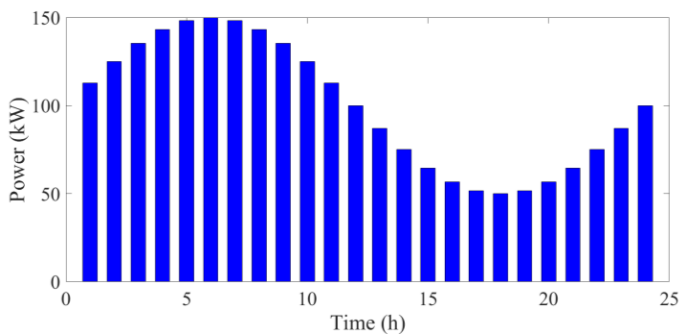

Fig. 2, Power demand for day-ahead in case 1.

Fig. 3 represents the optimal output power of DGs. Based on this figure, the cheapest unit (unit 1) is committed to generate power in many hours a day. Accordingly, unit 2 generates more power in comparison with unit 3. That means units' status is just based on the economic consideration which can demonstrates the effectiveness of the proposed algorithm.

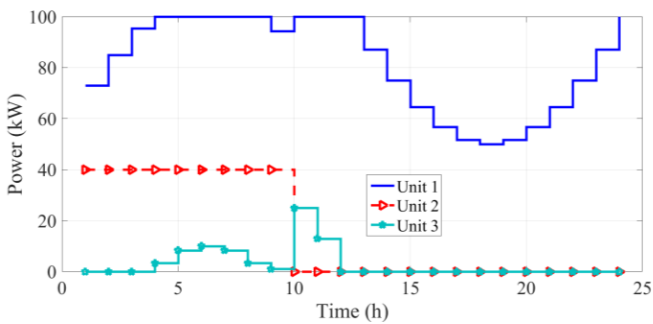

Fig. 3, Optimal output power of DGs in case1.

Moreover, the convergence speed of the proposed algorithm is compared with the particle swarm optimization (PSO) method as one of the well-known heuristics techniques in Fig. 4.

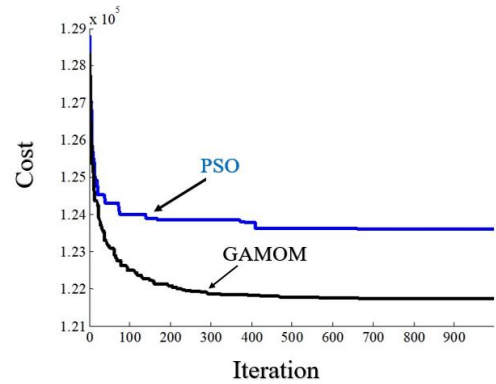

Fig. 4,Compare the convergence speed of GAMOM and PSO

Also, the total operation cost by the proposed algorithm is compared with other methods (heuristic and mathematical such as mixed integer linear programming (MILP)) in Table III.

Table III
Total Operation Costs

| Methods | Operation Cost Of Best Solution | Operation Cost Of Worst Solution |
|---|---|---|
| PSO | 36569.7557 | 38269.7557 |
| GA | 36269.7557 | 37369.7557 |
| GAMOM | 35279.7557 | 35289.7557 |
| MILP | 35269.7557 | - |

Simulation results and final operation cost of case 1 can prove the effectiveness of the proposed technique in comparison with other heuristic methods.

**Case2.** In this case, the optimal output powers of 10 generation units in the IEEE 39-bus test networked. Fig. 5 show the single line diagram of IEEE 39-bus test network which includes 10 generation units. Also, the total load demand of day-ahead is represents in Fig. 6. According to the simulation results, the there is no constraint violation that means the proposed algorithm is able to solve the complex problems as well.

Table IV presents the characteristics of generation units in case 2. According to the table, unit 1 the cheapest unit in comparison with others.

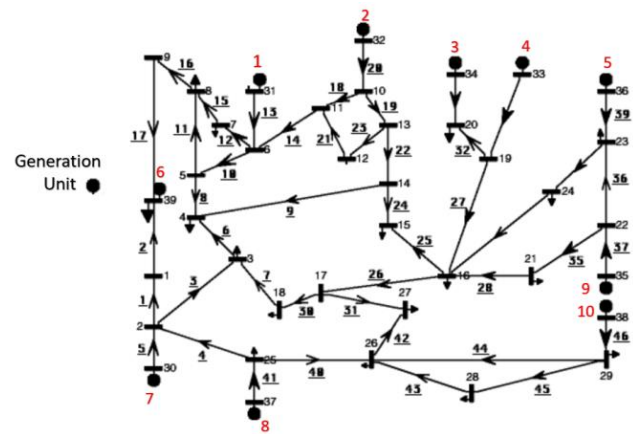

Fig. 5, IEEE 39 bus test network with 10 units

Table IV
Generation units charectristcis for case 2

|  | Unit 1 | Unit 2 | Unit 3 | Unit 4 | Unit 5 | Unit 6 | Unit 7 | Unit 8 | Unit 9 | Unit 10 |
|---|---|---|---|---|---|---|---|---|---|---|
| $a_i$ | 1000 | 970 | 700 | 680 | 450 | 370 | 480 | 660 | 665 | 670 |
| $b_i$ | 19.6 | 17.26 | 16.2 | 16.5 | 19.7 | 22.26 | 27.4 | 25.92 | 27.27 | 27.79 |
| $c_i$ | 0.00048 | 0.00031 | 0.002 | 0.00211 | 0.00398 | 0.00712 | 0.00079 | 0.00413 | 0.00222 | 0.00173 |
| $P_{max}$ | 455 | 455 | 130 | 130 | 162 | 80 | 85 | 55 | 55 | 55 |
| $P_{min}$ | 150 | 150 | 20 | 20 | 25 | 20 | 25 | 10 | 10 | 10 |
| r | 9.1 | 9.1 | 2.6 | 2.6 | 3.24 | 1.6 | 1.7 | 1.1 | 1.1 | 1.1 |
| RU | 227.5 | 227.5 | 65 | 65 | 81 | 40 | 42.5 | 27.5 | 27.5 | 27.5 |
| RD | 227.5 | 227.5 | 65 | 65 | 81 | 40 | 42.5 | 27.5 | 27.5 | 27.5 |
| $T_{ON}$ | 8 | 8 | 5 | 5 | 6 | 3 | 3 | 1 | 1 | 1 |
| $T_{OFF}$ | 8 | 8 | 5 | 5 | 6 | 3 | 3 | 1 | 1 | 1 |
| $T_{Cold}$ | 5 | 5 | 4 | 4 | 4 | 2 | 2 | 0 | 0 | 0 |
| $SU_{Cold}$ | 9000 | 9000 | 1100 | 1120 | 1800 | 340 | 520 | 60 | 60 | 60 |
| $SU_{Hot}$ | 4500 | 5000 | 550 | 560 | 900 | 170 | 260 | 30 | 30 | 30 |
| In.State | 8 | 8 | -5 | -5 | -6 | -3 | -3 | -1 | -1 | -1 |

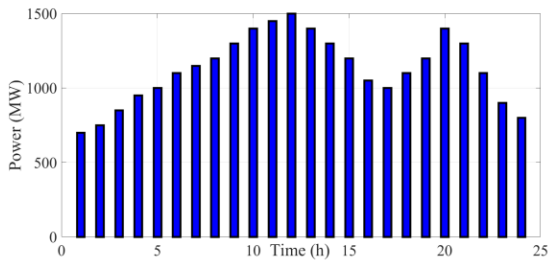

Fig. 6, Power demand for day-ahead in case 2.

Table V represents the status of generation units for the day-ahead. According to this table, where the status is one means the unit is committed and should be ON during this hour. Moreover, zero means the unit is OFF. For instance, unit one is ON entire the day. As another example, units 3 is OFF for 7 hours, after that it is committed for 14 hours, then turned OFF for 3 hours. The output power of generation units is represented in Fig. 7. It is worth noting that unit 1 is ON for the entire horizon with its maximum capacity due to its cheap price.

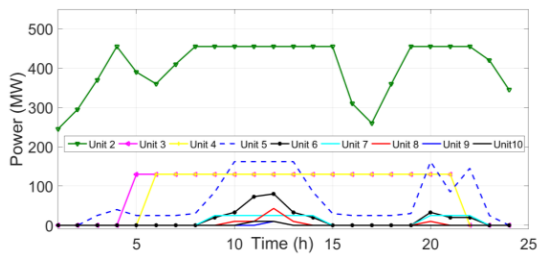

Fig. 7, Generation units output power in case 2.

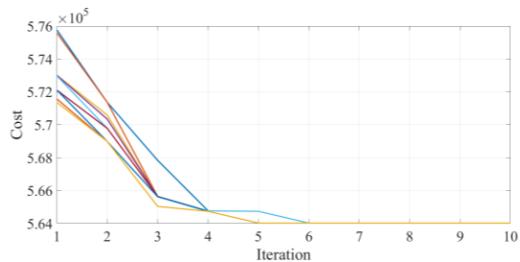

Fig. 9, Convergence speed of GAMOM in case 2.

Spinning reserve is one of most challenges in the SCUC problem and can has a significant effect on the cost function. Fig. 8, demonstrates the optimal output power of each generation unit as the spinning reserve. As it can see, the portion of the unit1 is more than others due to its cheapest price. Moreover, the convergence speed of the proposed algorithm show in Fig. 9. According to the figure, the optimal cost is obtained after just five iterations that can prove the high speed of the proposed algorithm. The total operation cost of this case is $555997.00.

Table V
Generation unit's status for the next 24 hours.

| Hour | Unit 1 | Unit 2 | Unit 3 | Unit 4 | Unit 5 | Unit 6 | Unit 7 | Unit 8 | Unit 9 | Unit 10 |
|---|---|---|---|---|---|---|---|---|---|---|
| 1 | 1 | 1 | 0 | 1 | 0 | 0 | 0 | 0 | 0 | 0 |
| 2 | 1 | 1 | 0 | 1 | 0 | 0 | 0 | 0 | 0 | 0 |
| 3 | 1 | 1 | 0 | 1 | 0 | 0 | 0 | 0 | 0 | 0 |
| 4 | 1 | 1 | 0 | 1 | 1 | 0 | 0 | 0 | 0 | 0 |
| 5 | 1 | 1 | 0 | 1 | 1 | 0 | 0 | 0 | 0 | 0 |
| 6 | 1 | 1 | 0 | 1 | 1 | 0 | 0 | 0 | 0 | 0 |
| 7 | 1 | 1 | 0 | 1 | 1 | 0 | 0 | 0 | 0 | 0 |
| 8 | 1 | 1 | 1 | 1 | 1 | 0 | 0 | 0 | 0 | 0 |
| 9 | 1 | 1 | 1 | 1 | 1 | 1 | 0 | 0 | 0 | 0 |
| 10 | 1 | 1 | 1 | 1 | 1 | 1 | 1 | 1 | 0 | 0 |
| 11 | 1 | 1 | 1 | 1 | 1 | 1 | 1 | 1 | 1 | 0 |
| 12 | 1 | 1 | 1 | 1 | 1 | 1 | 1 | 1 | 1 | 1 |
| 13 | 1 | 1 | 1 | 1 | 1 | 1 | 1 | 1 | 0 | 0 |
| 14 | 1 | 1 | 1 | 1 | 1 | 1 | 1 | 0 | 0 | 0 |
| 15 | 1 | 1 | 1 | 1 | 1 | 0 | 0 | 0 | 0 | 0 |
| 16 | 1 | 1 | 1 | 1 | 1 | 0 | 0 | 0 | 0 | 0 |
| 17 | 1 | 1 | 1 | 1 | 1 | 0 | 0 | 0 | 0 | 0 |
| 18 | 1 | 1 | 1 | 1 | 1 | 0 | 0 | 0 | 0 | 0 |
| 19 | 1 | 1 | 1 | 1 | 1 | 0 | 0 | 0 | 0 | 0 |
| 20 | 1 | 1 | 1 | 1 | 1 | 1 | 0 | 1 | 1 | 1 |
| 21 | 1 | 1 | 1 | 1 | 1 | 1 | 0 | 1 | 0 | 0 |
| 22 | 1 | 1 | 0 | 1 | 0 | 1 | 0 | 0 | 0 | 0 |
| 23 | 1 | 1 | 0 | 0 | 0 | 1 | 0 | 0 | 0 | 0 |
| 24 | 1 | 1 | 0 | 0 | 0 | 0 | 0 | 0 | 0 | 0 |

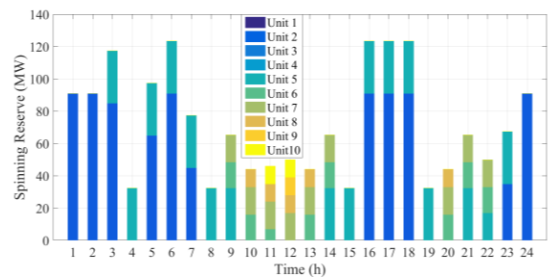

Fig. 8, Spinning reserve output power in case 2.

### V. Conclusion

In this paper, a novel modified GA based on GAMOM algorithm is applied to solve the SCUC problem. GAMOM is an evolutionary algorithm that includes two phases, GA phase and modified GA phase. The proposed algorithm is applied to two different cases as UC and SCUC problems. The results demonstrate the effectiveness and high performance of the proposed technique. Moreover, the convergence speed of the proposed algorithm is compared with PSO algorithm where it has a superiority as well.